\documentclass[runningheads]{llncs}
\usepackage[T1]{fontenc}
\usepackage{graphicx}
\usepackage{amsmath,amssymb}
\usepackage{booktabs,multirow}
\usepackage{makecell}
\usepackage{verbatim} 
\begin{document}
\title{FM-fMRI: Event Conditioned Flow Matching for Rest-to-Task fMRI Time-Series Synthesis}
\titlerunning{FM-fMRI: Flow Matching Task fMRI Time Series Synthesis}
%

\author{
Peiyu~Duan\inst{1},~
Jiyao~Wang\inst{1},~
Nicha~C.~Dvornek\inst{1,2},~\\
Junlin~Yang\inst{2},~
Ziqi~Gao\inst{1},~
Lawrence~H.~Staib\inst{1,2},~
James~S.~Duncan\inst{1,2.3}
}
\institute{Department of Biomedical Engineering, 
\and Department of Radiology \& Biomedical Imaging, \and Department of Electrical Engineering\\Yale University, New Haven, CT, USA \\
\email{{camille.duan}@yale.edu}
}
\authorrunning{P. Duan et al.}
  
\maketitle              
\begin{abstract}
Task-based fMRI provides a direct readout of task-evoked neural dynamics, but it is expensive and difficult to acquire at scale, motivating rest-to-task synthesis from widely available resting-state fMRI (rsfMRI). We propose \textbf{FM-fMRI}, an \textbf{event-conditioned flow-matching} model that learns a continuous-time conditional vector field to generate task ROI time series from a subject’s rsfMRI and the task event information. The formulation enables fast ODE-based sampling and flexible conditioning over heterogeneous event schedules. Rather than optimizing for pointwise reconstruction, we evaluated generated signals using complementary criteria that probe temporal and spectral structure, subject and group-level connectome consistency, and distributional alignment. On the public Human Connectome Project and internal Biopoint autism cohort, FM-fMRI achieves the strongest spectral and connectivity agreement and improved distribution-level matching over conditional diffusion, generative adversarial networks (GANs), and variational autoencoders (VAEs) baselines. Furthermore, we augment the BioPoint cohort by synthesizing task-fMRI ROI time series with our method, improving downstream autism classification and demonstrating practical utility in data-limited clinical settings. The code will be available on GitHub. 
\end{abstract}

\section{Introduction}
Task-based functional Magnetic Resonance Imaging (tfMRI) is a primary tool for quantifying how the brain reconfigures under controlled cognitive and affective demands, which can expose disorder-related alterations and compensatory dynamics, supporting more sensitive disease identification than resting state fMRI (rsfMRI) ~\cite{woo2017biomarkers}\cite{STNAGNN2024}. However, it is expensive and operationally fragile due to its strict experimental control, longer scan sessions, and sustained subject engagement. These constraints make tfMRI difficult to collect at scale and particularly challenging in pediatric, aging, and clinical cohorts. rsfMRI is comparatively easy to acquire and widely available, motivating rest-to-task generation that synthesizes task-evoked signals from rsfMRI to broaden access to task-like measurements and enable data augmentation when tfMRI is scarce.

Existing rest-to-task approaches often map resting connectivity to task activation or contrast maps, with recent graph neural networks, convolutional architectures, and transformers improving predictive performance \cite{tavor2016taskfree,jones2017resting,ngo2020connectomic,ngo2022brainsurfcNN,kwon2025swifun}. 
However, these formulations target static summaries rather than ROI-resolved task time series, and their deterministic objectives tend to regress toward conditional means, under-representing distributional variability and task-evoked dynamics. Generic multivariate time-series generators based on GANs, VAEs, and diffusion models can produce stochastic samples \cite{yoon2019timegan,yuan2024diffusionts,shen2024multires,liu2024retrievaldiff}, and neuroimaging-specific variants have begun to add task-alignment objectives for clinical prediction \cite{LiYif_Taskaligned_MICCAI2025}; yet they remain challenging to control with experimental schedules and are seldom constrained to preserve fMRI structure such as low-frequency spectra and functional connectivity. 
What is needed is a generative framework that natively supports structured conditioning, efficient sampling, and differentiable constraint enforcement. Flow matching and rectified flow address this by learning continuous-time vector fields for conditional generation, enabling stable training and ODE-based sampling with a direct handle for incorporating structured, differentiable constraints \cite{lipman2022flowmatching,liu2022rectifiedflow,lipman2024fmguide}. Despite these advantages, flow-based formulations have rarely been explored for rest-conditioned, timing-controlled tfMRI time-series synthesis with explicit spectral and connectomic constraints.

We propose \textbf{FM-fMRI}, an \textbf{event-conditioned flow-matching} framework that synthesizes task-evoked ROI time series from a subject's rsfMRI and experimental schedule timing via a learned continuous-time conditional vector field. Our main contributions are threefold: (1) an event-conditioned flow-matching formulation for rest-to-task fMRI generation; (2) evaluation prioritizing neurobiological realism, such as spectral, connectomic, and distribution-level validity, including recovery of group-level connectivity structure, over pointwise reconstruction; (3) strong synthesis performance across multiple tfMRI tasks with synthesized task signals improving downstream learning in data-limited clinical settings.

\section{Method}
\begin{figure}
    \centering
    \includegraphics[width=\linewidth]{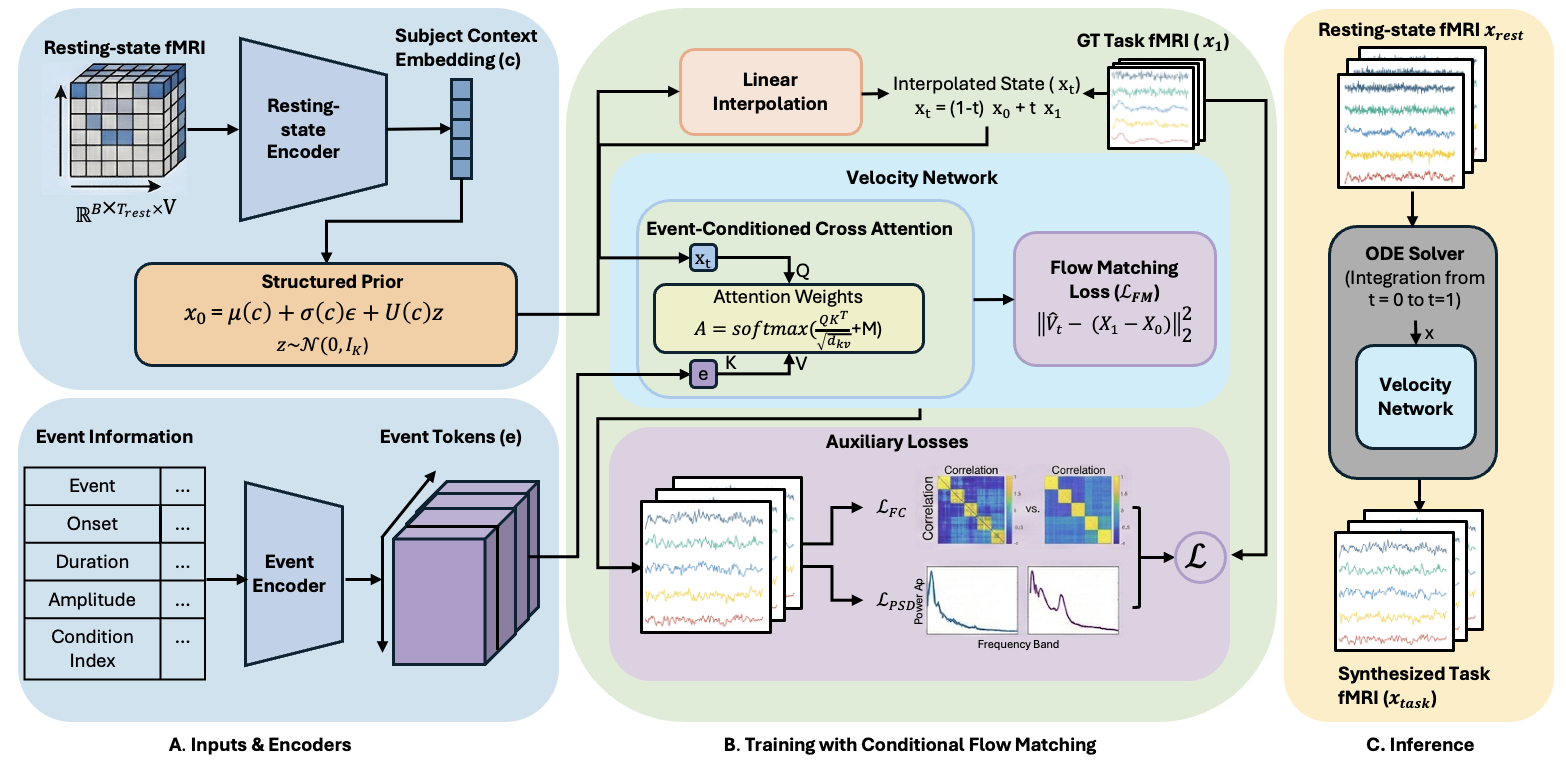}
    \caption{\textbf{FM-fMRI overview.} (A) Inputs and encoders: resting-state ROI time series are encoded to parameterize a structured prior, while event information is encoded into event tokens; (B) Training: we learn an event-conditioned velocity field via cross-attention, optimized with flow-matching, connectivity and spectrum-aware objectives. (C) Inference: task time series trajectories are synthesized by integrating the learned ODE.}
    \label{fig:architecture}
\end{figure}

We model rest-to-task synthesis as the conditional generation of task-evoked ROI time series given resting-state dynamics and experimental timing (Fig.~\ref{fig:architecture}), $x_{\text{task}} \sim p_\theta(x_{\text{task}} \mid x_{\text{rest}},e )$
where $x_{\text{task}} \in \mathbb{R}^{T \times V}$ denotes ROI time series with $V$ regions and $T$ time points, $x_{\text{rest}}$ is resting-state input, and $e$ encodes task event information. 

\textbf{Resting-state encoder} Given resting-state input $x_{\text{rest}} \in \mathbb{R}^{T_{\text{rest}} \times V}$, we extract a subject-specific context embedding $c = f_{\text{enc}}(x_{\text{rest}})$ using a patch-based Transformer encoder. A prepended \texttt{[CLS]} token aggregates sequence information, and its final representation is used as the resting context. 

\textbf{Structured prior}
Rather than using a typical flow-matching setup initialized from isotropic Gaussian noise, we learn a subject context-dependent structured prior specific to the fMRI signal:
\begin{equation}
x_0 = \mu(c) + \sigma(c)\,\epsilon_{\text{colored}} + U(c) z\quad z\sim \mathcal{N}(0, I_K),
\end{equation}
where $\mu(c)$ and $\sigma(c)$ are rest-conditioned per-ROI means and scale heads, and $U(c)\in\mathbb{R}^{V\times K}$ is a rest-conditioned low-rank spatial factor with rank $K=8$ that captures structured cross-ROI covariance. The term $\epsilon_{\text{colored}}$ is a temporal template with the power spectral density proportional to $1/f$, to better capture the elevated low-frequency power of fMRI signals~\cite{biswal1995fc}, and stabilize training.

\textbf{Task fMRI Event token}
Task event information is derived from FSL-style event timing files \cite{jenkinson2012fsl}, which contain information regarding the types of tasks performed $s$, the time duration $d$, stimulus amplitude $a$, and task onset $o$. For the $k$-th event, we parse the information into tuples $(o_k,d_k,a_k,s_k)$. We convert $o_k,d_k$ to TR units (divided by TR) and $z$-score $a_k$. The task event token (e) in Fig.\ref{fig:architecture} is embedded as
\begin{equation}
e_k=\phi_{\text{MLP}}([o_k,d_k,a_k]) + E_{\text{cond}}(s_k),
\end{equation}
where $\phi_{\text{MLP}}$ projects continuous timing features and $E_{\text{cond}}$ is a learned embedding for the task event type.

\textbf{Event-conditioned cross-attention}
To allow time-specific modulation by experimental structure, event tokens are incorporated through cross-attention within the velocity network~\cite{vaswani2017attention}. Queries are computed from the linear projection of the current state, $Q = W_q x_t$, while keys and values are derived from event tokens $K = W_k e$, $V = W_v e$. Attention weights are
$A = \mathrm{softmax}\!\left( \frac{QK^\top}{\sqrt{d_{\text{ev}}}} + M \right)$,
where $M$ masks padded events. 
The resulting event context $\mathrm{e}_{\text{ctx}} = A V$ is concatenated with $x_t$, the resting context $c$, and the time embedding before predicting the velocity. This enables each time point to attend selectively to relevant task events.

\textbf{Conditional flow matching}
We train a conditional velocity field $v_\theta(t, x_t, c, e)$ using flow matching. Given prior samples $x_0$ and real task signals $x_1$, we construct linear interpolants $x_t = (1-t)x_0 + t x_1, \quad t \sim \mathcal{U}(0,1)$
with target velocity $v^*(x_t, t) = x_1 - x_0$. The flow-matching objective is
\begin{equation}
\mathcal{L}_{\text{FM}}
=
\mathbb{E}_{t, x_0}
\left[
\left\|
v_\theta(t, x_t, c, e)
-
(x_1 - x_0)
\right\|_2^2
\right].
\end{equation}
In practice, $v_\theta$ is parameterized by a lightweight MLP applied pointwise over time: we concatenate $x_t$, the resting context $c$, a learned time embedding $\psi(t)$, and an event-derived context vector.

\textbf{Connectivity and spectrum-aware objectives} We add auxiliary losses to preserve neurophysiological structure beyond pointwise fidelity. For functional connectivity (FC), let $R_{ij}(x)$ denote the Pearson correlation between ROI $i$ and $j$; we define a weighted FC loss which emphasizes preservation of strong task-relevant connections.
\begin{equation}
\mathcal{L}_{\text{FC}}
=
\sum_{i<j}
w_{ij}\big(R_{ij}(\hat{x}_1) - R_{ij}(x_1)\big)^2,
\quad
w_{ij} = |R_{ij}(x_1)|^2,
\end{equation}
 For spectral fidelity, let $P_i(f)=|\mathcal{F}(x_i)(f)|^2$ denote the power spectrum of ROI $i$; over a physiologically relevant band $\mathcal{B}$ (0.01--0.05~Hz)~\cite{zuo2010oscillating}, we use
\begin{equation}
\mathcal{L}_{\text{PSD}}
=
\sum_{i}\sum_{f \in \mathcal{B}}
\left(\log P_i^{\hat{x}_1}(f)-\log P_i^{x_1}(f)\right)^2.
\end{equation}
The overall training objective is
$
\mathcal{L}
=
\mathcal{L}_{\text{FM}}
+
\lambda_{\text{FC}}\mathcal{L}_{\text{FC}}
+
\lambda_{\text{PSD}}\mathcal{L}_{\text{PSD}}.
$

We optimize all models with Adam (learning rate $1\times10^{-3}$; weight decay $1\times10^{-5}$) for 50 epochs and a batch size of 16. At test time, task trajectories are generated by integrating the learned ODE \begin{equation}\frac{d\mathbf{x}}{dt} = v_{\boldsymbol{\theta}}(t, x, c, e)
\end{equation}from $t=0$ to 1 using an explicit Euler fixed-step solver starting from the learned structured prior $x_0$, yielding $x(1)$ as the synthesized task fMRI signal \cite{chen2018neuralode}.

\section{Experiments and Results}

\textbf{Datasets and Data Preprocessing}
We evaluate rest-to-task synthesis with two cohorts that span large-scale normative fMRI and a clinical task-fMRI setting. 
For the Human Connectome Project (HCP) \cite{VanEssen2012HCPAcquisition}, we use 1{,}025 subjects with paired rsfMRI and tfMRI across seven paradigms.  For each run, we extract regional mean BOLD time series using the AAL atlas \cite{aal_atlas}, yielding paired resting inputs and task targets. 
For FM-fMRI, we additionally leverage the task timing files, which contain stimulus onsets, duration, and conditioning information. To assess generalization beyond HCP, we also evaluate on the Biopoint cohort \cite{Kaiser2010NeuralSignaturesAutism} (118 participants; 75 with autism and 43 controls), extracting ROI time series with the Shen268 atlas \cite{shen268}. 
For both datasets, we use subject-disjoint splits of 70\% for training, 15\% for validation, 15\% for test to prevent data leakage.

\begin{table}[t]
\centering
\caption{Subject-level generation performance grouped by metric. Best values are bolded.}
\label{tab:baseline_by_metric}
\scriptsize
\setlength{\tabcolsep}{1pt} 
\renewcommand{\arraystretch}{0.95} 
\begin{tabular}{@{}llccccccc@{}}
\toprule
\textbf{Metric} & \textbf{Model} & \textbf{Emotion} & \textbf{Gambling} & \textbf{Language} & \textbf{Motor} & \textbf{Relational} & \textbf{Social} & \textbf{WM} \\
\midrule

\multirow{6}{*}{PSD $\downarrow$} 
 & TimeGAN\cite{yoon2019timegan} & 2.0546 & 1.9923 & 1.9109 & 1.9390 & 1.9305 & 1.9179 & 1.9371 \\
 & TimeVAE\cite{sohn2015cvae} & 2.0517 & 1.9721 & 1.8791 & 1.8829 & 1.8805 & 1.8560 & 1.9093 \\
 & Diffusion-TS\cite{yuan2024diffusionts} & 4.0047 & 4.0942 & 3.9848 & 4.0620 & 3.9983 & 4.1018 & 3.9994 \\
& DDPM\cite{ho2020ddpm} & 1.7793 & 1.6921 & 1.5944 & 1.6350 & 1.6719 & 1.7894 & 1.4799 \\
& LSTM-GAN\cite{zhu2019lstmagan} & 2.1006 & 2.0959 & 1.9916 & 1.9727 & 2.0687 & 2.1236 & 1.8706\\
 &\textbf{FM-fMRI} & \textbf{1.6001} & \textbf{1.5399} & \textbf{1.4222} & \textbf{1.3634} & \textbf{1.4305} & \textbf{1.5601} & \textbf{1.3189} \\
\midrule
\multirow{6}{*}{FC sim. $\uparrow$} 
 & TimeGAN\cite{yoon2019timegan} & 0.2546 & 0.2229 & 0.2307 & 0.1723 & 0.2473 & 0.2630 & 0.2296 \\
 & TimeVAE\cite{sohn2015cvae} & 0.2795 & 0.2653 & 0.2640 & 0.2433 & 0.2942 & 0.3078 & 0.2580 \\
 & Diffusion-TS\cite{yuan2024diffusionts}& 0.0442 & 0.0168 & 0.0109 & 0.0026 & 0.0123 & 0.0078 & 0.0073 \\
 & DDPM\cite{ho2020ddpm} & 0.2677 & 0.2618 & 0.2454 & 0.2429 & 0.2056 & 0.2087 & 0.2274 \\
& LSTM-GAN\cite{zhu2019lstmagan} & 0.2935 & 0.1432 & 0.2305 & 0.2439 & 0.2593 & 0.3399 & 0.1968 \\
 & \textbf{FM-fMRI} & \textbf{0.5684} & \textbf{0.5455} & \textbf{0.5272} & \textbf{0.5302} & \textbf{0.5606} & \textbf{0.6046} & \textbf{0.5366} \\
\midrule
\multirow{6}{*}{cFID $\downarrow$} 
 & TimeGAN\cite{yoon2019timegan} & 98.06 & 114.91 & 127.31 & 151.15 & 121.47 & 120.66 & 116.97 \\
 & TimeVAE\cite{sohn2015cvae} & 105.62 & 116.22 & 129.48 & 154.58 & 127.17 & 123.59 & 118.43 \\
 & Diffusion-TS\cite{yuan2024diffusionts} & 177.12 & 188.78 & 194.03 & 220.74 & 194.29 & 202.80 & 189.29 \\
& DDPM\cite{ho2020ddpm} & 35.69 & 41.16 & 39.59 & 44.66 & 40.43 & 46.80 & 48.78 \\
& LSTM-GAN\cite{zhu2019lstmagan} & 113.76 & 257.37 & 128.54 & 142.30 & 159.14 & 102.28 & 184.56 \\
 & \textbf{FM-fMRI}& \textbf{24.1311} & \textbf{30.6712} & \textbf{37.5067} & \textbf{35.5717} & \textbf{29.8250} & \textbf{30.7688} & \textbf{39.9683} \\
 \midrule
 \multirow{6}{*}{MAE $\downarrow$} 
 & TimeGAN\cite{yoon2019timegan} & 0.7676 & 0.7658 & 0.7662 & 0.7702 & 0.7551 & 0.7527 & 0.7670 \\
 & TimeVAE\cite{sohn2015cvae} & \textbf{0.7661} & \textbf{0.7620} & \textbf{0.7562} & \textbf{0.7532} & \textbf{0.7471} & \textbf{0.7431} & \textbf{0.7529} \\
 & Diffusion-TS\cite{yuan2024diffusionts} & 0.9393 & 0.9382 & 0.9315 & 0.9445 & 0.9351 & 0.9392 & 0.9174 \\
 & DDPM\cite{ho2020ddpm} & 0.8016 & 0.8005 & 0.8040 & 0.8019 & 0.8002 & 0.8029 & 0.8025 \\
 & LSTM-GAN\cite{zhu2019lstmagan} & 0.7797 & 0.7904 & 0.7877 & 0.7785 & 0.7888 & 0.7785 & 0.7913 \\
 & \textbf{FM-fMRI} & 0.9342 & 0.9413 & 0.9508 & 0.9599 & 0.9588 & 0.9569 & 0.9437 \\

\midrule
\multirow{6}{*}{P@5\% $\uparrow$} 
& TimeGAN\cite{yoon2019timegan} & 0.2173 & 0.2186 & 0.2289 & 0.1775 & 0.2986 & 0.1731 & 0.1809 \\
& TimeVAE\cite{sohn2015cvae} & 0.2215 & 0.2130 & 0.2329 & 0.2157 & 0.3010 & 0.2046 & 0.1904 \\
 & Diffusion-TS\cite{yuan2024diffusionts} & 0.1599 & 0.2426 & 0.2318 & 0.2361 & 0.2175 & 0.1748 & 0.3084 \\
& DDPM\cite{ho2020ddpm} & 0.2641 & 0.2506 & 0.2137 & 0.2210 & 0.2181 & 0.2208 & 0.2635 \\
 & LSTM-GAN\cite{zhu2019lstmagan} & 0.2394 & 0.1231 & 0.1240 & 0.1836 & 0.1485 & 0.2342 & 0.1082 \\
 & \textbf{FM-fMRI} & \textbf{0.4617} & \textbf{0.4329} & \textbf{0.4352} & \textbf{0.4404} & \textbf{0.5000} & \textbf{0.4825} & \textbf{0.4535} \\
\bottomrule
\end{tabular}
\end{table}


\subsection{Baseline and Metrics}
We benchmark FM-fMRI against conditional DDPM~\cite{ho2020ddpm}, Diffusion-TS~\cite{yuan2024diffusionts}, TimeVAE~\cite{kingma2013vae,sohn2015cvae}, TimeGAN~\cite{yoon2019timegan}, and LSTM-GAN~\cite{zhu2019lstmagan}, all conditioned on a resting-state context embedding. We use tfMRI-aligned metrics:PSD~\cite{welch1967welch} discrepancy to assess preservation of the spectrum, FC similarity calculated by the Pearson correlation of the FC map, top-5\% edge recovery (P@5\%) to evaluate connectome structure, and cFID~\cite{heusel2017fid} to quantify conditional distributional alignment in FC space. Since rest-to-task synthesis is many-to-many, we prioritize second-order structure and report MAE only as an amplitude sanity check.

\begin{figure}[t]
  \centering
  \includegraphics[width=\linewidth]{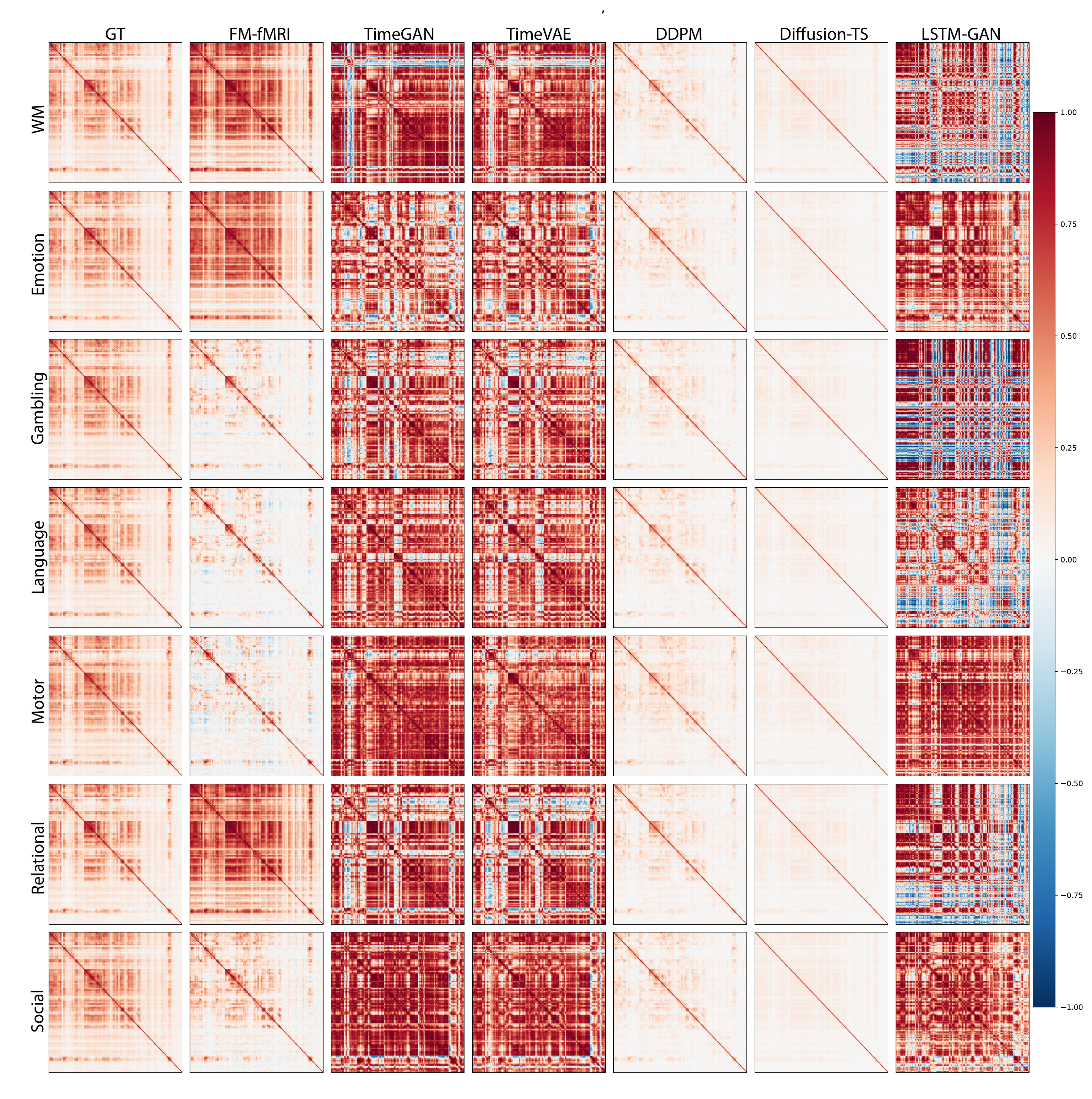}
  \caption{Group-level FC comparison on all HCP tasks.}
  \label{fig:group_fc_comparison_emotion}
\end{figure}

\textbf{Task fMRI Synthesis on HCP and Biopoint}
HCP results in Table~\ref{tab:baseline_by_metric} highlight the distinction between pointwise reconstruction fidelity and neurophysiological realism in rest-to-task synthesis. Methods such as TimeVAE achieve the lowest MAE, consistent with variance-shrinking conditional-mean predictions that match average trajectories yet over-smooth task-evoked variability and distort spectral or connectomic structure shown in Fig.~\ref{fig:group_fc_comparison_emotion}. FM-fMRI exhibits higher MAE than reconstruction-focused baselines, which is expected given that it is not optimized for strict pointwise waveform matching. Instead, FM-fMRI preserves realistic low-frequency power and cross-ROI covariance geometry, leading to improved PSD and FC realism and stronger conditional population alignment, which better reflects downstream connectivity and group-level analyses.

On Biopoint (Table~\ref{tab:biopoint_wide}), despite cohort and atlas differences, the same pattern holds: baseline rankings vary across MAE and PSD, whereas FM-fMRI consistently yields the strongest connectivity realism and distributional matching, supporting robustness under dataset and atlas shifts.

\begin{table}[t]
\centering
\caption{Biopoint subject-level generation performance. Best values are bolded.}
\label{tab:biopoint_wide}
\scriptsize
\setlength{\tabcolsep}{5pt} 
\begin{tabular}{lccccc}
\toprule
\textbf{Model} & \textbf{MAE $\downarrow$} & \textbf{PSD $\downarrow$} & \textbf{FC sim. $\uparrow$} & \textbf{FC P@5\% $\uparrow$} & \textbf{cFID $\downarrow$} \\
\midrule
TimeGAN\cite{yoon2019timegan}        & 0.7667 & 2.5819 & 0.1672 & 0.0977 & 168.76 \\
TimeVAE\cite{sohn2015cvae}        & \textbf{0.7659} & 2.5791 & 0.2679 & 0.1688 & 118.60 \\
Diffusion-TS\cite{yuan2024diffusionts}   & 1.0650 & 2.6324 & 0.0940 & 0.0855 & 116.99 \\
DDPM\cite{ho2020ddpm}           & 0.8218 & 2.3643 & 0.1402 & 0.1344 & 104.53 \\
LSTM-GAN\cite{zhu2019lstmagan}       & 0.7672 & 2.5852 & 0.0012 & 0.0532 & 455.66 \\
\textbf{FM-fMRI}      & 0.7923 & \textbf{2.3445} & \textbf{0.3508} & \textbf{0.2436} & \textbf{43.32} \\
\bottomrule
\end{tabular}
\end{table}

\subsection{Ablation: Event Conditioning and Auxiliary Objectives}

Table~\ref{tab:grouped_tasks_ev_aux_boldbest_check} ablates event token and auxiliary connectivity (Aux) losses on HCP. The full model with event information input and connectivity objective is the most reliable configuration, achieving the best or near-best performance across tasks on structure-aware metrics, indicating that task-timing cues and explicit neurophysiological regularization are jointly necessary for realistic synthesis. The ablation also highlights complementary effects: Aux alone drives substantial gains in spectral and connectivity fidelity relative to the baseline, whereas event information alone often improves pointwise alignment but does not consistently enhance FC-space distributional matching. Together, these results reinforce that MAE is insufficient to characterize rest-to-task realism and that combining event token conditioning with spectrum/FC-aware objectives yields the most consistent improvements on metrics aligned with downstream neuroimaging analyses.

\begin{table}[t]
\centering
\caption{Ablation results on Event information and Aux losses. Best values are bolded.}
\label{tab:grouped_tasks_ev_aux_boldbest_check}
\scriptsize
\setlength{\tabcolsep}{4.5pt} 
\begin{tabular}{lccccccc}
\toprule
\textbf{Task} & \textbf{Event} & \textbf{Aux} & \textbf{MAE} $\downarrow$ & \textbf{PSD} $\downarrow$ & \textbf{FC Sim} $\uparrow$ & \textbf{P@5} $\uparrow$ & \textbf{cFID} $\downarrow$ \\
\midrule

\multirow{4}{*}{WM} & -- & -- & 0.8625 & 2.0286 & 0.2665 & 0.2528 & 44.1144 \\
 & -- & \checkmark & 0.9425 & 1.3688 & 0.5025& 0.4111& 41.8149 \\
 & \checkmark & -- & \textbf{0.8549} & 2.0158 & 0.3183 & 0.2682 & 64.9692 \\
 & \checkmark & \checkmark & 0.9437 & \textbf{1.3189} & \textbf{0.5366} & \textbf{0.4535} & \textbf{39.9683} \\
\midrule
\multirow{4}{*}{Emotion} & -- & -- & \textbf{0.8184} & 2.0238 & 0.3391 & 0.2879 & 33.0135 \\
 & -- & \checkmark & 0.9436 & 1.6475 & 0.5095 & 0.4077 & 44.4805 \\
 & \checkmark & -- & 0.8448 & 1.9540 & 0.4219 & 0.2977 & 65.4704 \\
 & \checkmark & \checkmark & 0.9342 & \textbf{1.6001} & \textbf{0.5684} & \textbf{0.4617} & \textbf{24.1311} \\
\midrule

\multirow{4}{*}{Gambling} & -- & -- & 0.8297 & 1.9546 & 0.2472 & 0.2391 & 47.6530 \\
 & -- & \checkmark & 0.9382 & 1.6074 & 0.5039 & 0.3949 & 46.1616 \\
 & \checkmark & -- & \textbf{0.8262} & 1.9416 & 0.4720 & 0.3613 & 37.1983 \\
  & \checkmark & \checkmark & 0.9413 & \textbf{1.5399} & \textbf{0.5455} & \textbf{0.4329} & \textbf{30.6712} \\
\midrule

\multirow{4}{*}{Language} & -- & -- & 0.8535 & 1.9730 & 0.2875 & 0.2669 & 38.2063 \\
 & -- & \checkmark & 0.9602 & 1.4749 & 0.5007 & 0.3944 & 60.30 \\
 & \checkmark & -- & \textbf{0.8505} & 1.9669 & 0.3543 & 0.2867 & 59.6663 \\
  & \checkmark & \checkmark & 0.9508 & \textbf{1.4222} & \textbf{0.5272} & \textbf{0.4352} & \textbf{37.5067} \\
\midrule

\multirow{4}{*}{Motor} & -- & -- & 0.8805 & 2.1643 & 0.2380 & 0.2408 & 50.4327 \\
 & -- & \checkmark & 0.9604 & 1.7153 & 0.4886 & 0.4124 & 39.5460 \\
 & \checkmark & -- & \textbf{0.8748} & 2.1286 & 0.2966 & 0.2576 & 67.4445 \\
 & \checkmark & \checkmark & 0.9599 & \textbf{1.3634} & \textbf{0.5302} & \textbf{0.4404} & \textbf{35.5717} \\

\midrule

\multirow{4}{*}{Relational} & -- & -- & 0.8544 & 1.9700 & 0.2986 & 0.2733 & 40.0110 \\
 & -- & \checkmark & 0.9539 & 1.4997 & 0.5064 & 0.4559 & 45.1027 \\
 & \checkmark & -- & \textbf{0.8516} & 1.9615 & 0.3614 & 0.2894 & 61.1435 \\
 & \checkmark & \checkmark & 0.9588 & \textbf{1.4305} & \textbf{0.5606} & \textbf{0.5000} & \textbf{29.8250} \\
\midrule

\multirow{4}{*}{Social} & -- & -- & \textbf{0.8588} & 2.0207 & 0.3265 & 0.2718 & 64.3218 \\
 & -- & \checkmark & 0.9763 & 1.6057 & 0.5568 & 0.4313 & 47.2015 \\
 & \checkmark & -- & 0.8639 & 2.0199 & 0.2685 & 0.2539 & 43.3593 \\

 & \checkmark & \checkmark & 0.9569 & \textbf{1.5601} & \textbf{0.6046} & \textbf{0.4825} & \textbf{30.7688} \\
\bottomrule

\end{tabular}
\end{table}

\subsection{Downstream Classification Augmentation on Biopoint}

We assess downstream utility on Biopoint autism classification with three graph-based models under three training regimes: rsfMRI-only, tfMRI-only without augmentation, and tfMRI augmented with our proposed model. Synthetic tfMRI is generated for all biopoint cohort in the training split using its rsfMRI without access to validation/test subjects. The final performance is reported on a held-out test set (Table~\ref{tab:biopoint_cls_combined}). tfMRI-only generally surpasses rsfMRI-only performance, implying added discriminative information in task-evoked signals, except for the lowest-capacity classifier. Across all classifiers and metrics, augmentation with our synthesized signals yields consistent performance gains, with the largest improvements for the most state-of-the-art model (STAGIN) \cite{kim2021stagin}. These results demonstrate that generative rest-to-task modeling provides a practical and reliable augmentation strategy for data-scarce clinical cohorts.

\begin{table}[ht!]
\centering
\caption{Biopoint ASD classification performance under three settings: rest-only, task-based with and w/o augmentation. Best performance is bolded.}
\label{tab:biopoint_cls_combined}
\small
\setlength{\tabcolsep}{3.5pt}
\resizebox{\columnwidth}{!}{%
\begin{tabular}{lcccccccccccc}
\toprule
\multirow{2}{*}{\textbf{Model}}
& \multicolumn{4}{c}{\textbf{Rest-only}}
& \multicolumn{4}{c}{\textbf{Task w/o Augmentation}}
& \multicolumn{4}{c}{\textbf{Task with Augmentation}} \\
\cmidrule(lr){2-5} \cmidrule(lr){6-9} \cmidrule(lr){10-13}
& \textbf{Acc.} $\uparrow$ & \textbf{F1} $\uparrow$ & \textbf{AUC} $\uparrow$ & \textbf{Sens.} $\uparrow$
& \textbf{Acc.} $\uparrow$ & \textbf{F1} $\uparrow$ & \textbf{AUC} $\uparrow$ & \textbf{Sens.} $\uparrow$
& \textbf{Acc.} $\uparrow$ & \textbf{F1} $\uparrow$ & \textbf{AUC} $\uparrow$ & \textbf{Sens.} $\uparrow$ \\
\midrule
GCN\cite{GCN}
& 0.4583 & 0.5185 & 0.5333 & 0.4667
& 0.4583 & 0.5517 & 0.4963 & 0.5333
& \textbf{0.6250} & \textbf{0.6897} & \textbf{0.6741} & \textbf{0.6667} \\
GAT\cite{GAT}
& 0.5417 & 0.5926 & 0.6074 & 0.5333
& 0.5833 & 0.6154 & 0.6074 & 0.5333
& \textbf{0.6250} & \textbf{0.6400} & \textbf{0.7259} & \textbf{0.5333} \\
STAGIN\cite{kim2021stagin}
& 0.5833 & 0.6875 & 0.5037 & 0.7333
& 0.6667 & 0.7333 & 0.6741 & 0.7333
& \textbf{0.7500} & \textbf{0.7609} & \textbf{0.7481} & \textbf{0.8667} \\
\bottomrule
\end{tabular}%
}
\end{table}

\section{Conclusion and Discussion}
We presented \textbf{FM-fMRI}, an event-conditioned flow-matching model for rest-to-task fMRI time-series synthesis. Across seven HCP tasks, FM-fMRI delivers the best spectral fidelity, functional-connectivity recovery, and FC-based distributional alignment, underscoring the importance of structure-aware evaluation beyond pointwise reconstruction. Ablations confirm that event conditioning and connectivity regularization are both necessary for distribution-level performance. On the data-limited Biopoint autism cohort, synthesized task-like signals consistently improve downstream autism classification, with the largest gains for STAGIN. FM-fMRI remains effective across cohorts and atlas choices. Future work will extend conditioning and test cross-site robustness and calibration.

\bibliographystyle{splncs04}
\bibliography{reference}
\end{document}